\documentclass{article}
\usepackage{spconf,amsmath,graphicx}
\usepackage{amsfonts}
\usepackage[table,xcdraw]{xcolor}
\usepackage[]{algorithm2e}
\usepackage{multirow}
\usepackage{placeins}
\usepackage{comment}
\usepackage{url}
\usepackage{cite}
\usepackage{pgfplots}
\usepackage{siunitx}
\pgfplotsset{compat=1.9}
\usepgfplotslibrary{statistics}
\usetikzlibrary{pgfplots.groupplots}
\usetikzlibrary{patterns,positioning,fadings,through}
\usepackage{amsfonts}
\usepackage{multirow}
\usepackage{siunitx}
\usepackage{adjustbox}
\usepackage{balance}
\usepackage{makecell}
\usepackage{array}
\newcolumntype{?}{!{\vrule width 1.8pt}}
\newcommand{\PreserveBackslash}[1]{\let\temp=\\#1\let\\=\temp}
\newcolumntype{C}[1]{>{\PreserveBackslash\centering}p{#1}}
\newcolumntype{R}[1]{>{\PreserveBackslash\raggedleft}p{#1}}
\newcolumntype{L}[1]{>{\PreserveBackslash\raggedright}p{#1}}

\title{On the reversibility of adversarial attacks}

\name{Chau Yi Li*, Ricardo S\'{a}nchez-Matilla, Ali Shahin Shamsabadi, Riccardo Mazzon, Andrea Cavallaro\thanks{We thank the Alan Turing Institute (EP/N510129/1), which is funded by the U.K. Engineering
and Physical Sciences Research Council, for its support through the project PRIMULA. \newline*Contact author: chauyi.li@qmul.ac.uk}}
\address{Centre for Intelligent Sensing, Queen Mary University of London, UK}
\begin{document}
\ninept
\maketitle

\begin{abstract}
Adversarial attacks modify images with perturbations that  change the prediction of classifiers. These modified images, known as adversarial examples, expose the vulnerabilities of deep neural network classifiers. In this paper, we investigate the predictability of the mapping between the classes predicted for original images and for their corresponding adversarial examples. This predictability relates to the possibility of retrieving the original predictions and hence reversing the induced misclassification. We refer to this property as the reversibility of an adversarial attack, and quantify reversibility as the accuracy in retrieving the original class or the true class of an adversarial example.  We present an approach that reverses the effect of an adversarial attack on a classifier using a prior set of classification results. We analyse the reversibility of state-of-the-art adversarial attacks on benchmark classifiers and discuss the factors that affect the reversibility.
\end{abstract}


\section{Introduction}
\label{sec:intro}

Deep Neural Networks (DNNs) for visual classification tasks achieve high average accuracy in predicting the {true class} labels in benchmark datasets~\cite{he2016deep,krizhevsky2012imagenet}. However, DNNs are vulnerable to perturbations that modify an original image to mislead the classifier~\cite{szegedy2013intriguing, goodfellow2014}. These modified images, known as adversarial examples, facilitate the study of the robustness of DNN classifiers~\cite{tramer2017space}. Adversarial examples have also been used to conceal private information in images by precluding a classifier from identifying specific class labels~\cite{li2019scene}. 

Studies of adversarial examples analysed the relationship between the classes predicted for the adversarial examples (\textit{adversarial classes}) and the corresponding \textit{true classes} of the original images~\cite{abbasi2017robustness}, and between the adversarial classes and the classes predicted for the original images (\textit{original classes})~\cite{li2019scene}. However, the possibility of retrieving the original predictions from the adversarial class, a property which we refer to as {\em reversibility}, has so far been neglected. 

In this paper, we argue that reversibility is a fundamental property of an adversarial attack, and propose to quantify the extent to which an attack may be reversed. We achieve this goal by analysing the {mapping} between the predicted classes of original images and those of adversarial examples generated by an attack. We discuss the reversibility of state-of-the-art attacks on benchmark classifiers and investigate the factors affecting the reversibility. Moreover, we present an approach to perform {Prediction Reversal from the Adversarial Class} (PRAC) using the prediction results of a set of original images and their corresponding adversarial classes. Specifically, for each adversarial class, we obtain the probabilities that its original images would be predicted to be of a particular original class. Our previous work on reversibility analysed the dissimilarity between the frequency distribution of adversarial classes and a uniform distribution, without  attempting to retrieve the original class~\cite{li2019scene}. 

PRAC is useful for designers of adversarial attacks and defences. In fact, PRAC allows an attack designer to quantify the extent to which the attack can be reversed, whereas a defence designer may label a set of adversarial examples and use PRAC to retrieve the original and true class of new adversarial examples generated with the same attack. 



\section{Class mapping}
\label{sec: proposed}

Let~$C(\cdot)$ be a~$D$-class DNN classifier that predicts the class of an image to be one in~\mbox{$\{y_1, \ldots, y_i, \ldots, y_D\}$}. Let the output of an original image~$x$ passing through all but the last layer of the classifier,~$L(\cdot)$, be defined as
\begin{equation}
  L(x) = \Big(
  p(y_1 | x), p(y_2 | x), \cdots , p(y_D | x)\Big),
  \label{eq: classifier_confidence}
\end{equation}
where~\mbox{$p(y_k|x)$} is the probability of~$x$ being of class~$y_k$. Let the original class,~\mbox{$y_i=C(x)$}, be the classifier's output, \mbox{where~$i$ is defined by}
\begin{equation}
  i = \underset{k \in \{1,\ldots,D\} }{\arg\max} p(y_k | x).
\end{equation}
When the classification is inaccurate, the {original class} differs from the {\em true class} of the image.

Let~$\dot{x}$ be the adversarial example generated from~$x$, such that the {\em adversarial class},~\mbox{$y_j = C(\dot{x})$}, differs from the {\em original class}, i.e.~\mbox{$C(\dot{x}) \neq C(x)$}. Targeted attacks  choose a specific adversarial class~\cite{li2019scene,papernot2016limitations,carlini2017towards,tramer2017ensemble,kurakin2016adversarialscale,kurakin2016adversarial,xie2018improving}, whereas {non-targeted} attacks only aim to mislead the classifier~\cite{kurakin2016adversarial,xie2018improving,MoosaviDezfooli16,modas2018sparsefool}. Adversarial attacks may also specify that the probability of the {adversarial class},~\mbox{$p(y_j|\dot{x})$}, should exceed a pre-defined value,~$\sigma$~\cite{kurakin2016adversarial}.

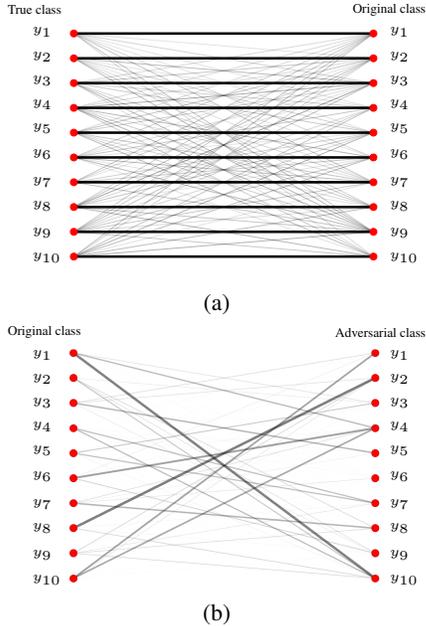
\begin{figure}[t]
\centering
\begin{tabular}{c}
\begin{tikzpicture}
\node (MNIST_NFGSM) at (0,0) {\includegraphics[width=0.53\columnwidth]{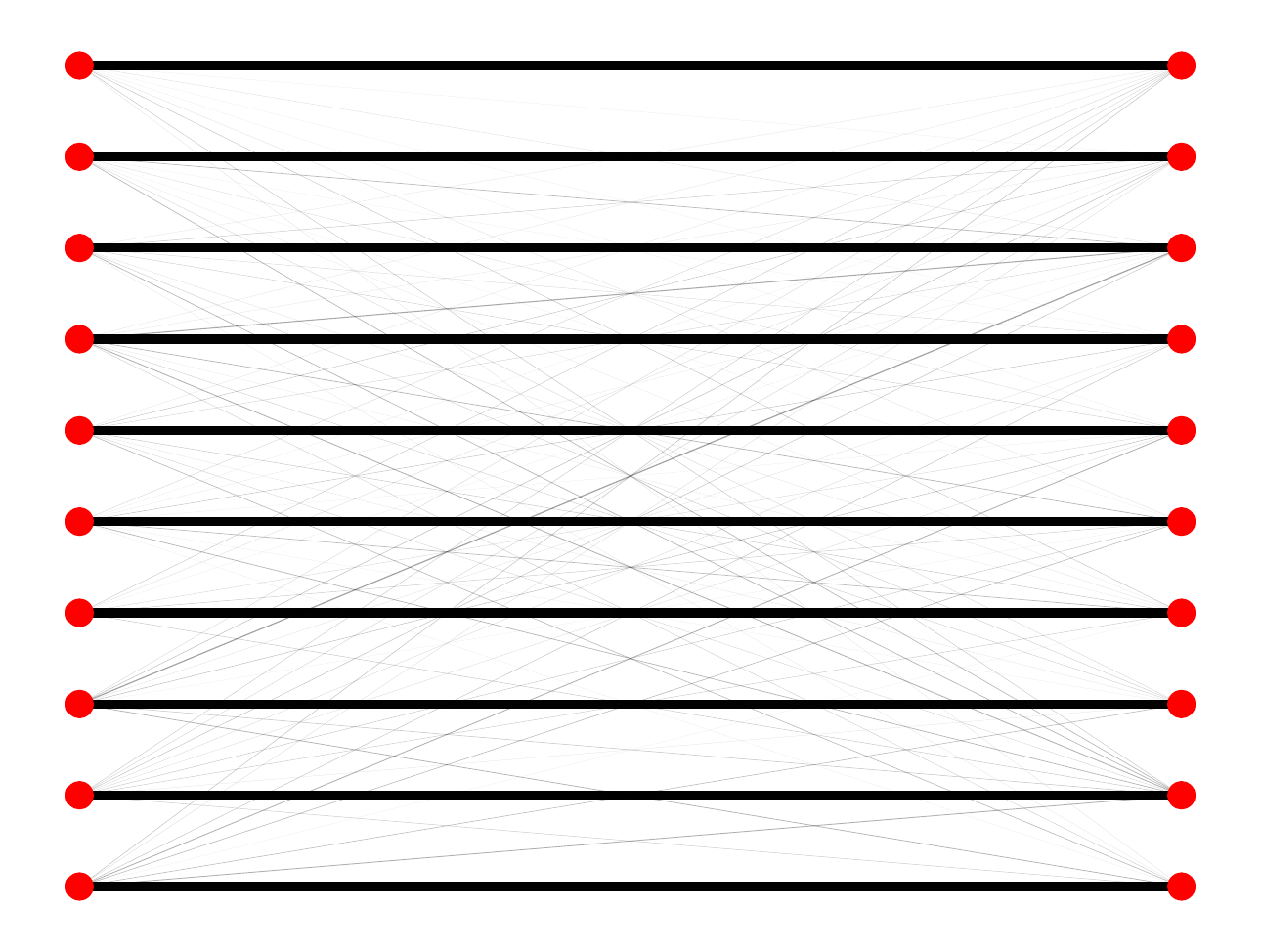}};
\node [anchor=east] () at (2.8,1.8) {\tiny Original class};
\node [anchor=west] () at (2.05,1.5) {\tiny~$y_1$};
\node [anchor=west] () at (2.05,1.18) {\tiny~$y_2$};
\node [anchor=west] () at (2.05,0.86) {\tiny~$y_3$};
\node [anchor=west] () at (2.05,0.54) {\tiny~$y_4$};
\node [anchor=west] () at (2.05,0.22) {\tiny~$y_5$};
\node [anchor=west] () at (2.05,-0.12) {\tiny~$y_6$};
\node [anchor=west] () at (2.05,-0.49) {\tiny~$y_7$};
\node [anchor=west] () at (2.05,-0.81) {\tiny~$y_8$};
\node [anchor=west] () at (2.05,-1.16) {\tiny~$y_9$};
\node [anchor=west] () at (2.05,-1.5) {\tiny~$y_{10}$};
\node [anchor=west] () at (-2.98,1.8) {\tiny True class};
\node [anchor=west] () at (-2.7,1.5) {\tiny~$y_1$};
\node [anchor=west] () at (-2.7,1.18) {\tiny~$y_2$};
\node [anchor=west] () at (-2.7,0.86) {\tiny~$y_3$};
\node [anchor=west] () at (-2.7,0.54) {\tiny~$y_4$};
\node [anchor=west] () at (-2.7,0.22) {\tiny~$y_5$};
\node [anchor=west] () at (-2.7,-0.12) {\tiny~$y_6$};
\node [anchor=west] () at (-2.7,-0.49) {\tiny~$y_7$};
\node [anchor=west] () at (-2.7,-0.81) {\tiny~$y_8$};
\node [anchor=west] () at (-2.7,-1.16) {\tiny~$y_9$};
\node [anchor=west] () at (-2.7,-1.5) {\tiny~$y_{10}$};
\end{tikzpicture}\\
(a) \\
\begin{tikzpicture}
\node (MNIST_NFGSM) at (0,0) {\includegraphics[width=0.492\columnwidth]{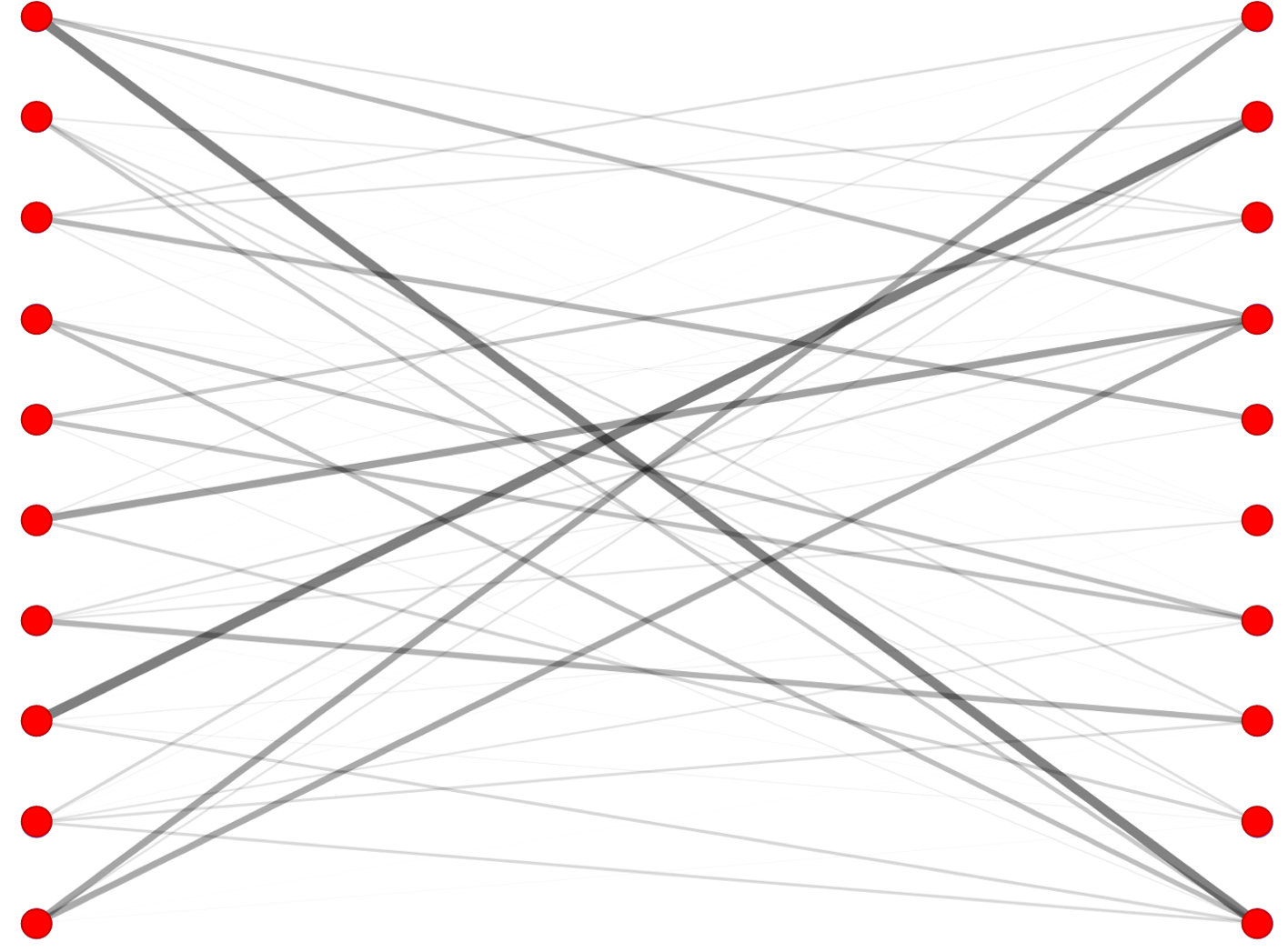}};
\node [anchor=east] () at (2.8,1.8) {\tiny Adversarial class};
\node [anchor=west] () at (2.05,1.5) {\tiny~$y_1$};
\node [anchor=west] () at (2.05,1.18) {\tiny~$y_2$};
\node [anchor=west] () at (2.05,0.86) {\tiny~$y_3$};
\node [anchor=west] () at (2.05,0.54) {\tiny~$y_4$};
\node [anchor=west] () at (2.05,0.22) {\tiny~$y_5$};
\node [anchor=west] () at (2.05,-0.12) {\tiny~$y_6$};
\node [anchor=west] () at (2.05,-0.49) {\tiny~$y_7$};
\node [anchor=west] () at (2.05,-0.81) {\tiny~$y_8$};
\node [anchor=west] () at (2.05,-1.16) {\tiny~$y_9$};
\node [anchor=west] () at (2.05,-1.5) {\tiny~$y_{10}$};
\node [anchor=west] () at (-2.98,1.8) {\tiny Original class};
\node [anchor=west] () at (-2.7,1.5) {\tiny~$y_1$};
\node [anchor=west] () at (-2.7,1.18) {\tiny~$y_2$};
\node [anchor=west] () at (-2.7,0.86) {\tiny~$y_3$};
\node [anchor=west] () at (-2.7,0.54) {\tiny~$y_4$};
\node [anchor=west] () at (-2.7,0.22) {\tiny~$y_5$};
\node [anchor=west] () at (-2.7,-0.12) {\tiny~$y_6$};
\node [anchor=west] () at (-2.7,-0.49) {\tiny~$y_7$};
\node [anchor=west] () at (-2.7,-0.81) {\tiny~$y_8$};
\node [anchor=west] () at (-2.7,-1.16) {\tiny~$y_9$};
\node [anchor=west] () at (-2.7,-1.5) {\tiny~$y_{10}$};
\end{tikzpicture}\\
(b) \\
\end{tabular}
\caption{Mapping between the source class (left nodes) and destination class (right nodes). The thickness (and opacity) of an edge is proportional to the mapping frequency between the two classes. (a) Mapping from the true class to the original class predicted by LeNet classifier on the MNIST~\cite{lecun-mnisthandwrittendigit} training dataset. The classifier has an accuracy of 97.35\% on the original images, therefore the horizontal edges dominate. (b) Mapping from the original class to the adversarial class induced by the non-targeted attack~\mbox{N-FGSM}~\cite{kurakin2016adversarial}. Note that the mapping is not uniform.
}
\label{fig:mapping_graph_bipartite}
\end{figure}

To investigate the predictability of mapping between the predicted classes, we visualise the {mapping} as a weighted bipartite graph, where the nodes are class labels (the same class labels aligned horizontally) and the edges are weighed by the mapping frequency between the classes. The weight of an edge determines its thickness (and opacity): a thicker (and darker) edge indicates a more frequent mapping between the classes. Fig.~\ref{fig:mapping_graph_bipartite} shows two examples. Fig.~\ref{fig:mapping_graph_bipartite}(a) visualises the class mapping from the true class to the original class of the MNIST training dataset predicted by LeNet classifier~\cite{lecun-mnisthandwrittendigit}. The training dataset contains 60,000 images, distributed unevenly among the true classes\footnote{There are between 5,421 and 6,742 instances in each true class.
}. LeNet has a classification accuracy  of 97.35\% on the original images, resulting in the dominating thick horizontal edges. The thin~\mbox{non-horizontal} edges identify the 2.65\% inaccurate classifications.  
Fig.~\ref{fig:mapping_graph_bipartite}(b) visualises the class mapping for the adversarial examples generated by a non-targeted FGSM attack~(N-FGSM)~\cite{kurakin2016adversarial} for the same classifier and dataset. Note that certain class pairs have thicker edges that indicate a preferential mapping. For example, most images with adversarial class label~$y_{10}$ originate from class~$y_{1}$. The existence of such preferential mapping indicates the possibility of reversing the adversarial attack and hence retrieving the original class.

\section{Reversibility}
\label{sec: proposed_measure}

We aim to quantify the reversibility of an adversarial attack, which measures the extent to which the original predictions can be retrieved from the adversarial class. To this end, we propose an approach that performs {Prediction Reversal from an Adversarial Class} (PRAC) by determining the probability that the adversarial class is from a certain original class. PRAC requires the classifier,~$C(\cdot)$; a set~$\mathcal{S}$ of original images~$x$ with original predicted classes covering the~$D$ possible classes; and the adversarial classes~\mbox{$y_j=C(\dot{x})$} of the corresponding adversarial examples~\mbox{$\dot{x}$ for all~\mbox{$x\in \mathcal{S}$}}. 

PRAC encodes the probability of the adversarial class~\mbox{$y_j=C(\dot{x})$} being from any of the~$D$ original classes in a vector
\begin{equation}
    \mathbf{p}_j = (p_j^1, p_j^2,\cdots, p_j^D), 
\end{equation} 
where~$p_j^k$ is the probability that the original images are predicted to be of the original class~$y_k$. Each~$p_j^k$ is defined as the average of~\mbox{$p(y_k|x)$} for all~\mbox{$x\in\mathcal{S}$}. This probability can be computed using all but the last layer of~$C(\cdot)$, with a prior expectation that the image has an equal probability of being of any one of the~$D$ original classes, i.e.~$1/D$, as
\begin{equation}
\label{eq:: probability_entry}
p_j^k = \frac{1}{1+\sum\limits_{i=1}^D n_{ij}} \Bigg(\frac{1}{D} + \sum_{\substack{
x \in \mathcal{S} \hspace{0.5em} \text{s.t.}\\ y_j = C(\dot{x})\\}} p(y_k|x)\Bigg),
\end{equation}
where~$n_{ij}$ is the number of images with original class~\mbox{$y_i=C(x)$} and adversarial class~\mbox{$y_j=C(\dot{x})$}. When none of the images in~$\mathcal{S}$ has an adversarial class~$y_j$ (hence~\mbox{$n_{ij} = 0$}), the probability~$p^k_j$ is~$1/D$.



When a new adversarial example~$\dot{x}$ generated by the same attack is classified as adversarial class~\mbox{$y_j=C(\dot{x})$}, PRAC reverses the prediction using~$\mathbf{p}_j$. Specifically, PRAC obtains the~\mbox{top-1} retrieved class as the class~$y_r$ with the highest probability in~$\mathbf{p}_j$, where
\begin{equation}
\label{eq: retrieve_class} 
 r = 
 \underset{k \in \{1,...,D\}}{\arg\max}~{p}_j^k. 
\end{equation}

We quantify reversibility as the~\textit{retrieval accuracy} of the retrieved class matching the original class or the true class. In particular, the~\mbox{top-$k$} retrieval accuracy,~$T_k$, can be computed from the total number of instances of the corresponding original or true class being in the~\mbox{top-$k$} retrieved classes,~$R_k$, as 
 \begin{equation}
 \label{eq: topk_accuracy}
{T_k} = 
  \frac{R_k}{N},
\end{equation}
where~$N$ is the total number of adversarial examples generated by the attack and not in the set~$\mathcal{S}$. A~\mbox{$T_5$} retrieval accuracy of 50\% of matching the original class means that the original class is one of the~\mbox{top-5} retrieved classes in half of the instances. 


\begin{table}[t]
\caption{Reversibility of N-FGSM~\cite{kurakin2016adversarial} attacking LeNet on the MNIST dataset~\cite{lecun-mnisthandwrittendigit}. The high original classification accuracy means only a few adversarial examples would select the true class of an image (\mbox{top-1} accuracy of 1.40\%). We report the retrieval {accuracy} of the~\mbox{top-1} retrieved class matching the original class and the true class using PRAC. }
\setlength\tabcolsep{2.8pt}
\centering
\small 
\vspace{.25cm}
\begin{tabular}{cS[table-format=3.2]C{1.3cm}C{1.3cm}rrrrrc}
\hline
 & {{\textbf{Classification}}} & \multicolumn{2}{c}{\textbf{Retrieved class}}  \\ 
 \cline{3-4}
 &  {\textbf{accuracy}} & \textit{original} & \textit{true} \\ 
 \hline 
\multicolumn{1}{l}{{Original}}	 & 97.35 & 
\multicolumn{2}{c}{\cellcolor[HTML]{C0C0C0}} \\ 
 \multicolumn{1}{l}{N-FGSM~\cite{kurakin2016adversarial}}	 & 1.40 
 & 40.15	 
 & 39.80	 
 \\ 
 \hline
\end{tabular}
\label{tbl: MNIST_demonstration}
\end{table}

\begin{figure*}[t!]
\centering
\setlength\tabcolsep{2pt}
\begin{tabular}{cccc}
 \includegraphics[width=0.492\columnwidth]{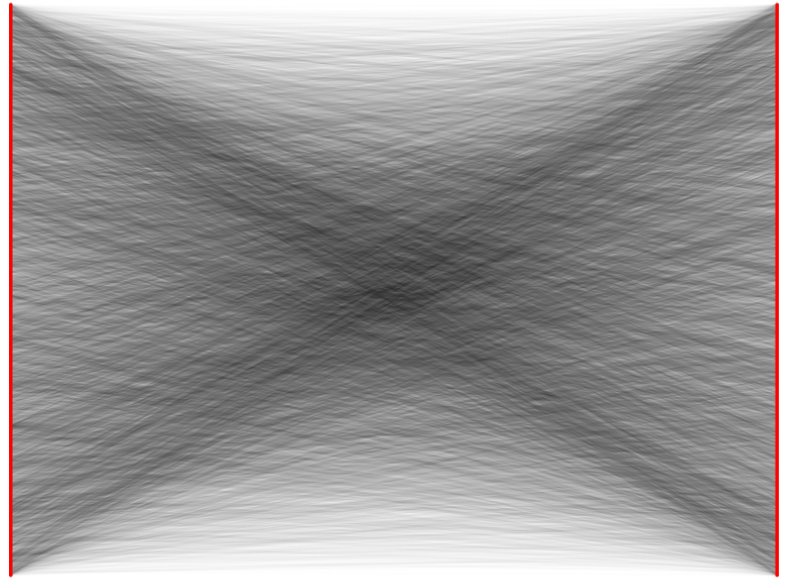}&
 \includegraphics[width=0.492\columnwidth]{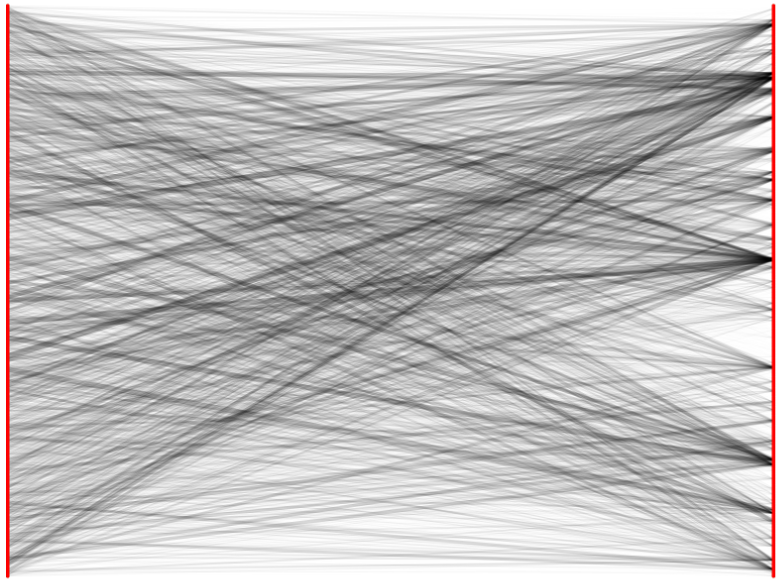}&
 \includegraphics[width=0.492\columnwidth]{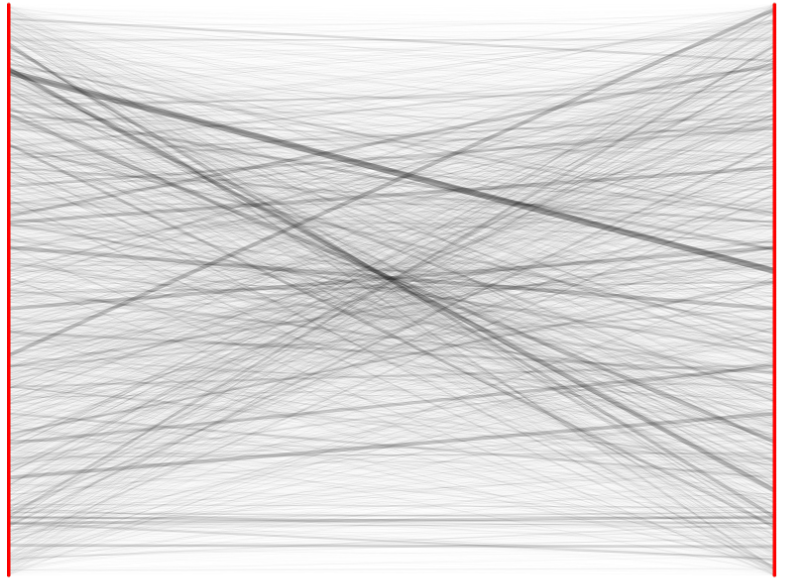} &
 \includegraphics[width=0.492\columnwidth]{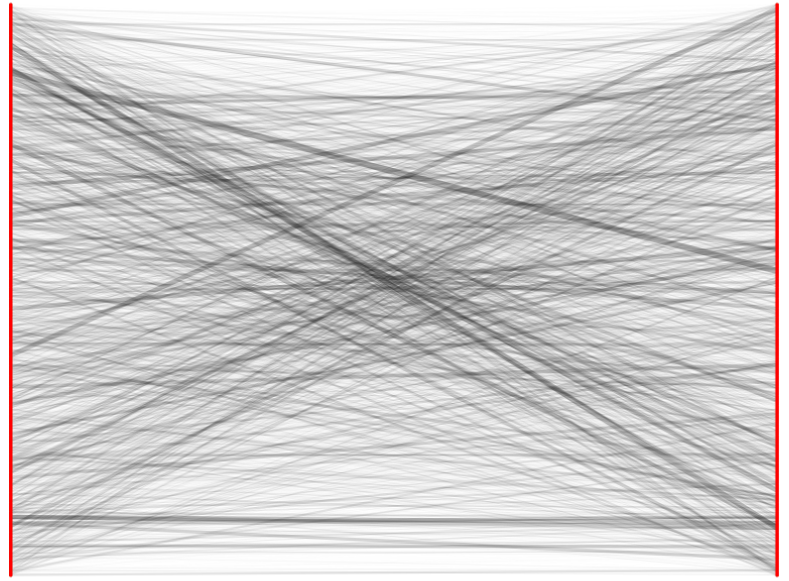}\\
 (a) R-FGSM & (b) L-FGSM & 
  (c) N-FGSM & (d) DeepFool\\
\end{tabular}
\caption{Class mapping for R-FGSM~\cite{kurakin2016adversarialscale}, L-FGSM~\cite{kurakin2016adversarial}, N-FGSM~\cite{kurakin2016adversarial} and DeepFool~\cite{MoosaviDezfooli16} attacking ResNet-50 classifier~\cite{he2016deep} on the Places365-Standard dataset~\cite{zhou2017places}. For each sub-image, the set of original classes is on the left, whereas the set of adversarial classes is on the right. 
 }
 \label{fig:mapping_graph_bipartite_places365}
\vspace{.2cm}
\end{figure*}


Tab.~\ref{tbl: MNIST_demonstration} shows the reversibility of N-FGSM attacking LeNet for the 10 MNIST classes. The probability vectors~$\mathbf{p}_j$ are constructed using the MNIST training set (see mapping in Fig.~\ref{fig:mapping_graph_bipartite}(b)), whereas the retrieval of the class is performed on the MNIST testing set which has 10,000 images. The probability of randomly selecting one class other than the adversarial class and matching the original class is~\mbox{1/ 9=11.11\%}, whereas the reversibility of N-FGSM is~40.15\%. The original high~\mbox{top-1} classification accuracy of the classifier~(97.35\%) results in the low~\mbox{top-1} classification accuracy of the adversarial examples generated by N-FGSM. In fact, the adversarial classes of 9,860 of the 10,000 examples (98.60\%) are different from their true classes. However, the~\mbox{top-1} class retrieved by PRAC matches the true class for 39.80\% of the images (3,980 images), which means that the true class of 40.40\% (3,980~divided by~9,860) of the adversarial examples, that concealed the true class, can be retrieved. 


%

%


\section{Results}
\label{sec: results}

\subsection{Set-up}
\label{subsec: set-up}

We analyse the reversibility of four widely used adversarial attacks: two targeted attacks, namely~\mbox{least-likely FGSM}~(L-FGSM)~\cite{kurakin2016adversarial} and targeted random FGSM (R-FGSM)~\cite{kurakin2016adversarialscale}; and two non-targeted attacks, namely non-targeted FGSM (N-FGSM)~\cite{kurakin2016adversarial} and DeepFool~\cite{MoosaviDezfooli16}. FGSM attacks generate adversarial examples that maximise the cost, computed by the loss function used in training, of remaining in the original class (non-targeted) or that minimise the cost of predicting the targeted class (targeted).~\mbox{L-FGSM} selects the least likely class from the original prediction as the target adversarial class, whereas R-FGSM performs the selection of the adversarial class with equal likelihood from all of the classes except the original class. DeepFool iteratively adds perturbations that orthogonally project the image to the nearest linearised class boundary (measured by the Euclidean distance), and eventually crossing the class boundary to induce misclassification. We re-implemented the FGSMs attacks and used the publicly available implementation of DeepFool. 

We use as classifiers AlexNet~\cite{krizhevsky2012imagenet},~\mbox{ResNet-18}~\cite{he2016deep} and ResNet-50~\cite{he2016deep} that are trained on the~\mbox{$D$=365} scene classes of the training set of the Places365-Standard dataset~\cite{zhou2017places}. Tab.~\ref{tbl:original_accuracy} shows the original classification accuracy of the three classifiers on the validation set:~AlexNet has the lowest accuracy and~\mbox{ResNet-50} is the most accurate classifier. Adversarial examples are crafted to change the original prediction of the classifier, thus more accurate classifiers label a larger number of adversarial images with a class that differs from the true class. We randomly split the validation set~\cite{zhou2017places} into two sets of equal size, one for constructing the probability vectors (Eq.~\ref{eq:: probability_entry}) and one for quantifying the reversibility of the attack (Eq.~\ref{eq: retrieve_class}). Each set has 18,250 images, evenly distributed among the 365 true classes. 


\begin{table}[t!]
\centering
\setlength\tabcolsep{1.8pt}
\caption{Classification accuracy (in percentage) on the original images of the Places-365 Standard validation dataset~\cite{zhou2017places}. 
} \label{tbl:original_accuracy}
\small \vspace{.25cm}
\begin{tabular}{lC{1.1cm}C{1.1cm}}
\hline
\multicolumn{1}{c}{{\textbf{DNN}}} 
 & \multicolumn{1}{c}{\mbox{\textbf{top-1}}} & \multicolumn{1}{c}{\mbox{\textbf{top-5}}} \\
 \hline
{AlexNet}~\cite{krizhevsky2012imagenet} & 46.76 & 77.66 \\ 
{\mbox{ResNet-18}}~\cite{he2016deep} & 52.67 & 83.28  \\ 
{\mbox{ResNet-50}}~\cite{he2016deep} & 54.25 & 84.98  \\
\hline
\end{tabular}
\end{table}
 
Fig.~\ref{fig:mapping_graph_bipartite_places365} visualises the mapping between original and adversarial classes for the attacks. R-FGSM is a reference of desirable behaviour as the target class is chosen with equal likelihood among all the possible classes. The strong preferential pattern of the targeted attack L-FGSM is caused by the target class selection procedure. The non-targeted attacks, N-FGSM and DeepFool, have similar patterns, due to the fact that the most probable adversarial classes of an original class label are limited to a few classes. To validate this, Tab.~\ref{tbl:matching_second_class} shows the percentage of adversarial classes matching any of the 2$^{\text{nd}}$ to 5$^{\text{th}}$ most likely classes in the original prediction. The adversarial class by a non-targeted attack often is the second most likely class in the original prediction (41.10\% and 65.79\% for N-FGSM and DeepFool attacking ResNet-50, respectively), followed by the 3$^{\text{rd}}$ and 4$^{\text{th}}$ most likely class. Despite the fact that the adversarial class can be any of the~\mbox{$D$-1=364} classes, at least 58.24\% adversarial classes of N-FGSM are in fact one of the 2$^{\text n\text d}$ to 5$^{\text{th}}$ most likely classes in the original prediction. The same observation can be made for DeepFool, with at least 92.10\% of the instances matching any of the top 2 to 5 classes. 
 

\subsection{Discussion}
\label{sec: about_the_proposed_method}

 \begin{table}[t]
\centering
\caption{Percentage of non-targeted adversarial classes matching the~$k^{\text{th}}$ most likely class in the original prediction on the Places-365 Standard dataset~\cite{zhou2017places}. }
\small \vspace{.25cm}
\setlength\tabcolsep{3pt}
\begin{tabular}{llccccc}
\hline
 \multicolumn{1}{c}{\multirow{2}{*}{\textbf{Attack}}} & \multicolumn{1}{c}{\multirow{2}{*}{\textbf{DNN}}} & 
 \multicolumn{1}{c}{\textbf{Any in}} & \multicolumn{4}{c}{\textbf{${k}^{\text{th}}$ most likely class}} \\ 
\cline{4-7}
& & \multicolumn{1}{c}{\textbf{2$^{\text{nd}}$ - 5$^{\text{th}}$}} & \multicolumn{1}{c}{{\hphantom{.}2$^{\text{nd}}$}} & \multicolumn{1}{c}{{\hphantom{.}3$^{\text{rd}}$}} & \multicolumn{1}{c}{\hphantom{.}{4$^{\text{th}}$}} & \multicolumn{1}{c}{\hphantom{.}{5$^{\text{th}}$}}\\
\hline
\multirow{3}{*}{N-FGSM~\cite{kurakin2016adversarial}} 
 &~\mbox{AlexNet}	 &  58.24 & 29.90 & 13.82 & 8.54 & 5.98 \\  
 &~\mbox{ResNet-18}	 & 70.38 & 37.85 & 16.68 & 9.35 & 6.50 \\
 &~\mbox{ResNet-50}	 & 73.87 &  41.10 & 17.02 & 9.61 & 6.14 \\  
 \hline 
 \multirow{3}{*}{DeepFool~\cite{MoosaviDezfooli16}} 
 &~\mbox{AlexNet}	 & 97.14 &  73.40 & 13.83 & 4.68 & 2.77 \\  
 &~\mbox{ResNet-18}	 & 94.74 &  67.11 & 17.11 & 5.26 & 5.26 \\ 
 &~\mbox{ResNet-50} & 92.10	 & 65.79 & 18.42 & 5.26 & 2.63 \\ 
 \hline
\end{tabular}
\label{tbl:matching_second_class}
\end{table}

We discuss the applicability of PRAC and analyse the reversibility of the attacks by reporting the accuracy of the~\mbox{top-1} and~\mbox{top-5} retrieved classes matching the original class and the true class.

We compare PRAC with a baseline frequency counting procedure that retrieves the top-$k$ classes as the~$k$ original classes that are most frequently mapped to the adversarial class. Tab.~\ref{tbl:proposed_original_groundtruth} compares the top-5 accuracy of PRAC and the baseline. We do not report the top-1 accuracy as there is little difference in the results. The baseline retrieved 11.90\% and 25.32\% true classes for L-FGSM and N-FGSM attacking ResNet-50, respectively, whereas PRAC retrieved 15.42\% and 32.82\%. Note that the baseline gives equal importance to all the original classes, regardless of whether they were accurately classified~(i.e. matching the true class),  whereas PRAC considers the predicted probabilities by the classifier~(Eq.~\ref{eq: retrieve_class}), which are often low when the classification is inaccurate. Therefore, PRAC gives more importance to the correctly classified instances and increases the probability of matching the true class. 


%
%
\begin{table}[t]
\centering\small 
\caption{Comparison between a frequency counting baseline and PRAC in terms of top-5 accuracy of the retrieved classes matching the original class and the true class.} 
\setlength\tabcolsep{2pt}
\vspace{.25cm}
\begin{tabular}{llC{1cm}C{1cm}@{\extracolsep{4pt}}C{1cm}C{1cm}}
\hline
\multicolumn{1}{c}{\multirow{2}{*}{\textbf{Attack}}} & \multicolumn{1}{c}{\multirow{2}{*}{\textbf{DNN}}} & \multicolumn{2}{c}{\textbf{Original class}} & \multicolumn{2}{c}{\textbf{True class}} \\
\cline{3-4}\cline{5-6}
 & & baseline & PRAC & baseline & PRAC \\
 \hline
\multirow{3}{*}{L-FGSM~\cite{kurakin2016adversarial}
} &~\mbox{AlexNet} &   17.92 & 17.92 & 11.36 & 15.38\\ 
&~\mbox{ResNet-18} &  17.92 & 18.20 & 11.90 & 15.62\\ 
&~\mbox{ResNet-50} &  17.92 & 18.22 & 11.90 & 15.42 \\ 
 \hline
 \multirow{3}{*}{N-FGSM~\cite{kurakin2016adversarial}} &~\mbox{AlexNet} & 37.65 & 37.34 & 25.12 & 34.43\\ 

 &~\mbox{ResNet-18} & 37.65 & 37.57 & 25.13 & 33.88\\ 
 &~\mbox{ResNet-50} &  37.65 & 37.70 & 25.32 & 32.82\\ 
 \hline
\end{tabular}
\label{tbl:proposed_original_groundtruth}
\end{table}

\begin{table}[t]
\centering
\caption{Classification accuracy on adversarial examples. Reversibility is quantified as the top-1 and top-5 retrieval accuracy,~$T_1$ and~$T_5$, of classes  retrieved with PRAC matching the original class or the true class. All reported values are in percentage.}
\small \vspace{.25cm}
\setlength\tabcolsep{1.6pt}
\begin{tabular}{llrr@{\extracolsep{4pt}}@{}S@{}@{}S@{}@{}S@{}@{}S@{}}
\hline
\multicolumn{1}{c}{\multirow{3}{*}{\textbf{Attack}}} & \multicolumn{1}{c}{\multirow{3}{*}{\textbf{DNN}}}& 
 \multicolumn{2}{c}{\multirow{1}{*}{\textbf{Classifier}}} & \multicolumn{4}{c}{\mbox{\textbf{Retrieved class}}} \\
\cline{5-8} 
& &  \multicolumn{2}{c}{\multirow{1}{*}{\textbf{accuracy}}} & \multicolumn{2}{c}{\mbox{\textit{original}}}  & \multicolumn{2}{c}{\mbox{\textit{true}}} \\
\cline{3-4}\cline{5-8} 
& &
\multicolumn{1}{r}{top-1} & \multicolumn{1}{r}{top-5} & \multicolumn{1}{c}{\hphantom{x}$T_1$} & \multicolumn{1}{c}{\hphantom{x}$T_5$} & \multicolumn{1}{c}{\hphantom{x}$T_1$} & \multicolumn{1}{c}{\hphantom{x}$T_5$} \\ 
\hline
 \multirow{3}{*}{R-FGSM~\cite{kurakin2016adversarialscale}} &~\mbox{AlexNet}	 &  	0.17 & 8.28  &  0.15 & 1.47 & 0.27	 & 1.44	 \\ 
 &~\mbox{ResNet-18}	 & 0.11	 & 6.88 & 0.27 & 1.44 & 0.29	 & 1.53	\\ 
  &~\mbox{ResNet-50} & 0.10 & 9.30 & 0.31 & 1.37 & 0.33 & 1.34 \\ 
\hline
 \multirow{3}{*}{L-FGSM~\cite{kurakin2016adversarial}} &~\mbox{AlexNet}	 & 0.03 & 0.61& 5.99 & 17.92	 & 5.00	 & 15.38 \\ 
 &~\mbox{ResNet-18}	 & 0.00	 & 0.41 & 6.13 & 18.20 & 5.15	 & 15.62 \\ 
 &~\mbox{ResNet-50} & 0.01 & 0.70 & 6.04 & 18.22 & 5.07 & 15.42 \\
\hline
 \multirow{3}{*}{N-FGSM~\cite{kurakin2016adversarial}} &~\mbox{AlexNet}	 & 14.25 & 52.60	 &  15.80 & 37.34 & 12.41	 & 34.43 \\ 
 &~\mbox{ResNet-18}	 & 14.52 & 56.05	 & 15.84 & 37.57 & 12.26	 & 33.88 \\ 
 &~\mbox{ResNet-50} & 16.72 & 64.06 & 16.30 & 37.70 & 12.52	 & 32.82\\
\hline
 \multirow{3}{*}{DeepFool~\cite{MoosaviDezfooli16}} &~\mbox{AlexNet} 	 & 33.41 & 74.61	 & 15.48 & 38.12 & 10.34	 & 30.63\\ 
 &~\mbox{ResNet-18}	 & 36.90& 79.95 & 17.52 & 44.02	  & 12.15	 & 35.63 \\ 
 &~\mbox{ResNet-50} & 42.54 & 82.59 & 18.68 & 44.48 & 13.15	 & 36.48 \\
 \hline
\end{tabular}
\label{tbl:results_places365_true_class}
\end{table}

Tab.~\ref{tbl:results_places365_true_class} shows the retrieval accuracy on the four adversarial attacks (Eq.~\ref{eq: retrieve_class}). The reversibility of an attack does not change significantly across the three  classifiers. This is expected as reversibility is a property of the attack, not the classifier. 
For non-targeted attacks, PRAC retrieves for N-FGSM on AlexNet the original class from the~\mbox{top-1} and~\mbox{top-5} 15.80\% and 37.34\% of the times, respectively. DeepFool is the most reversible attack as PRAC can retrieve the original class from the~\mbox{top-5} classifications with an {accuracy} of 38.12\% for AlexNet and at least 44.48\% for~\mbox{ResNet-18} and~\mbox{ResNet-50}. As for targeted attacks, R-FGSM, which randomly selects one out of the 364 classes as the target class, is the least reversible attack with a~\mbox{top-1} retrieval {accuracy} between 0.10\% and 0.17\% across classifiers. L-FGSM, which selects the least-likely original class as the target class, has the original class retrieved at least 5.99\% of the instances (i.e. the original class is retrieved for at least 1,108 adversarial examples). 

To analyse the possibility of the retrieved classes matching the true class, Tab.~\ref{tbl:results_places365_true_class} reports the accuracy of the true class matching the top-1 and the top-5 retrieved classes, and also the accuracy of the classifier on the adversarial examples. DeepFool has the smallest effect on classification accuracy and its adversarial effect is highly reversible. In particular, the retrieved class of DeepFool attacking ResNet-50 matches the true class at 36.48\% for top-5, which is equivalent to 42.90\% (36.48\% divided by 84.98\%) of the classifier's original accuracy (Tab.~\ref{tbl:original_accuracy}).~\mbox{L-FGSM} has the largest impact on classification accuracy, which varies between 0.00\% and 0.03\% for top-1, and between 0.41\% and 0.70\% for top-5. However, the retrieved class matches the true class with a relatively high accuracy (minimum~5\% and 15.38\% for top-1 and top-5, respectively). 
This large discrepancy between the low classification accuracy and top true class retrieval accuracy suggests that, when designing an adversarial attack, more objectives than just a low classification accuracy, should be considered to avoid the attack to be reversed. 

We finally investigate the effect of specifying the adversarial class probability on the reversibility of the non-targeted N-FGSM. Tab.~\ref{tbl:sigma_reversibility} shows the reversibility when the adversarial class probability is forced to exceed~\mbox{$\sigma\in\{0.6, 0.7, 0.8, 0.9\}$} for N-FGSM attacking AlexNet. We note that the average predicted probability of the adversarial classes without specifying the probability (i.e.~only aiming to mislead the classifier) is~0.48. The larger~$\sigma$, the less reversible the attack.  We believe that this happens because the preferential mapping that contributes to the high reversibility of non-targeted attacks  (Tab.~\ref{tbl:matching_second_class}) is weakened. In particular, as~$\sigma$ increases and becomes larger than the average probability of~0.48, the attack cannot attain such a high  probability with its original preferential class mapping and seeks another class as the adversarial one, resulting in a lower reversibility.

\begin{table}[t]
\centering
\setlength\tabcolsep{2.2pt}
\caption{Effect on the reversibility of specifying a minimum value~$\sigma$ for the adversarial class probability when using N-FGSM~\cite{kurakin2016adversarial} attacking AlexNet~\cite{krizhevsky2012imagenet} on the Places365-Standard dataset~\cite{zhou2017places}. N/A means that the only objective for the attack is to mislead the classifier, not to reach a specific probability for the adversarial class.
}\label{tbl:sigma_reversibility}
\small \vspace{.25cm}
\begin{tabular}{cR{1.3cm}R{1.3cm}rrrrrr}
\hline
 \multicolumn{1}{c}{\multirow{2}{*}{$\boldsymbol{\sigma}$}} & \multicolumn{2}{c}{\textbf{Retrieved original class}} \\ 
 \cline{2-3}
   & \multicolumn{1}{c}{\hphantom{xxxxx}$T_1$} & \multicolumn{1}{c}{\hphantom{xx}$T_5$}\\
 \hline
 N/A & 15.80	 & 37.34\\
.6	 	 &	10.34	 & 24.48\\
.7 & 9.52	 & 23.06	 \\
.8	 & 	8.77	 & 22.23	 \\
.9	 & 8.29	 & 21.59	\\
\hline
\end{tabular}
\end{table}

\section{Conclusion}
\label{sec: conclusion}

We discussed the reversibility of adversarial attacks, namely the possibility of retrieving the original class of an adversarial image and hence reversing the attack. We demonstrated how to perform {Prediction Reversal from the Adversarial Class} (PRAC) by exploiting the {mapping} between original and adversarial classes. PRAC supports the design of more effective adversarial attacks as well as the design of new defences. We quantified the reversibility as the accuracy of the original class being among the top-$k$ retrieved classes, and also investigated the possibility of retrieving the true class. Furthermore, we analysed the factors affecting the reversibility of an attack. Notably, the high reversibility of non-targeted attacks is caused by the high probability of the adversarial class being of one of the top most likely classes. Moreover, specifying a higher probability of the adversarial class reduces the reversibility of non-targeted attacks.

\clearpage

\bibliographystyle{IEEEbib}
\bibliography{egbib}
\end{document}